\documentclass[tinyml]{acmart}

\usepackage{algorithm}
\usepackage{algorithmic}
\usepackage{amsmath}
\usepackage{bm}
\usepackage{ulem} % for underline, etc
\usepackage{multirow} % for multirow
\usepackage{booktabs} % To thicken table lines
\usepackage{makecell}
\usepackage{graphicx}
\usepackage{color}
\usepackage{float}
\AtBeginDocument{%
  \providecommand\BibTeX{{%
    \normalfont B\kern-0.5em{\scshape i\kern-0.25em b}\kern-0.8em\TeX}}}

\setcopyright{rightsretained}
\copyrightyear{2022}
\acmYear{2022}

\begin{document}

\title{LDP: Learnable Dynamic Precision for Efficient Deep 
Neural Network Training and Inference}

\author{Zhongzhi Yu}
\email{zy42@rice.edu}
\affiliation{%
  \institution{Rice University}
  \streetaddress{6100 Main st.}
  \city{Houston}
  \state{Texas}
  \country{USA}
  \postcode{77005}
}

\author{Yonggan Fu}
\email{yf22@rice.edu}
\affiliation{%
  \institution{Rice University}
  \streetaddress{6100 Main st.}
  \city{Houston}
  \state{Texas}
  \country{USA}
  \postcode{77005}
}

\author{Shang Wu}
\email{sw99@rice.edu}
\affiliation{%
  \institution{Rice University}
  \streetaddress{6100 Main st.}
  \city{Houston}
  \state{Texas}
  \country{USA}
  \postcode{77005}
}

\author{Mengquan Li}
\email{ml121@rice.edu}
\affiliation{%
  \institution{Rice University}
  \streetaddress{6100 Main st.}
  \city{Houston}
  \state{Texas}
  \country{USA}
  \postcode{77005}
}

\author{Haoran You}
\email{hy34@rice.edu}
\affiliation{%
  \institution{Rice University}
  \streetaddress{6100 Main st.}
  \city{Houston}
  \state{Texas}
  \country{USA}
  \postcode{77005}
}

\author{Yingyan Lin}
\email{yingyan.lin@rice.edu}
\affiliation{%
  \institution{Rice University}
  \streetaddress{6100 Main st.}
  \city{Houston}
  \state{Texas}
  \country{USA}
  \postcode{77005}
}

% \affiliation{%

% \affiliation{%

% \affiliation{%

% \affiliation{%

% \affiliation{%

% \affiliation{%

% \affiliation{%


\begin{abstract}
Low precision deep neural network (DNN) training is one of the most effective techniques for boosting DNNs’ training efficiency, as it trims down the training cost from the finest bit level. While existing works mostly fix the model precision during the whole training process, a few pioneering works have shown that dynamic precision schedules help DNNs converge to a better accuracy while leading to a lower training cost than their static precision training counterparts. However, existing dynamic low precision training methods rely on manually designed precision schedules to achieve advantageous efficiency and accuracy trade-offs, limiting their more comprehensive practical applications and achievable performance.
To this end, we propose LDP, a \textbf{L}earnable \textbf{D}ynamic \textbf{P}recision DNN training framework that can automatically learn a temporally and spatially dynamic precision schedule during training towards optimal accuracy and efficiency trade-offs. It is worth noting that LDP-trained DNNs are by nature efficient during inference. Furthermore, we visualize the resulting temporal and spatial precision schedule and distribution of LDP trained DNNs on different tasks to better understand the corresponding DNNs’ characteristics at different training stages and DNN layers both during and after training, drawing insights for promoting further innovations. Extensive experiments and ablation studies (seven networks, five datasets, and three tasks) show that the proposed LDP consistently outperforms state-of-the-art (SOTA) low precision DNN training techniques 
in terms of
training efficiency and achieved accuracy trade-offs. For example, in addition to having the advantage of being automated, our LDP achieves a 0.31\% higher accuracy with a 39.1\% lower computational cost when training ResNet-20 on CIFAR-10 as compared with the best SOTA method.

\end{abstract}
\maketitle
\section{Introduction}
% The recent breakthroughs achieved by deep neural networks (DNNs) rely on massive training data and huge model sizes, imposing prohibitive training costs that have raised environmental concerns and are at odds with the growing demand for on-device training to maintain the model accuracy with dynamic real-world data. For trimming down the training cost, 
% one of the most promising approaches is the low precision training, which adopts a low precision for the weights, activations, and gradients during training~\cite{banner2018scalable,sun2019hybrid,zhou2016dorefa} to reduce the training cost at the most fine-grained granularity. Additionally, their resulting DNNs have a lower inference cost than their floating-point counterparts by nature.
The recent breakthroughs achieved by deep neural networks (DNNs) rely on massive training data and huge model sizes, imposing prohibitive training costs that have raised environmental concerns and standing at odds with the growing demand for on-device training to maintain the model accuracy under dynamic real-world environments. For trimming down the training cost, 
one of the most promising approaches is low precision training, which adopts a precision lower than 32-bit floating-point for model weights, activations, and gradients during training~\cite{banner2018scalable,sun2019hybrid,zhou2016dorefa} to reduce the training cost at the most fine-grained granularity. Additionally, their resulting DNNs by nature have a lower inference cost than their floating-point counterparts.

\begin{figure}
    \centering
    \resizebox{\linewidth}{!}{
    \includegraphics{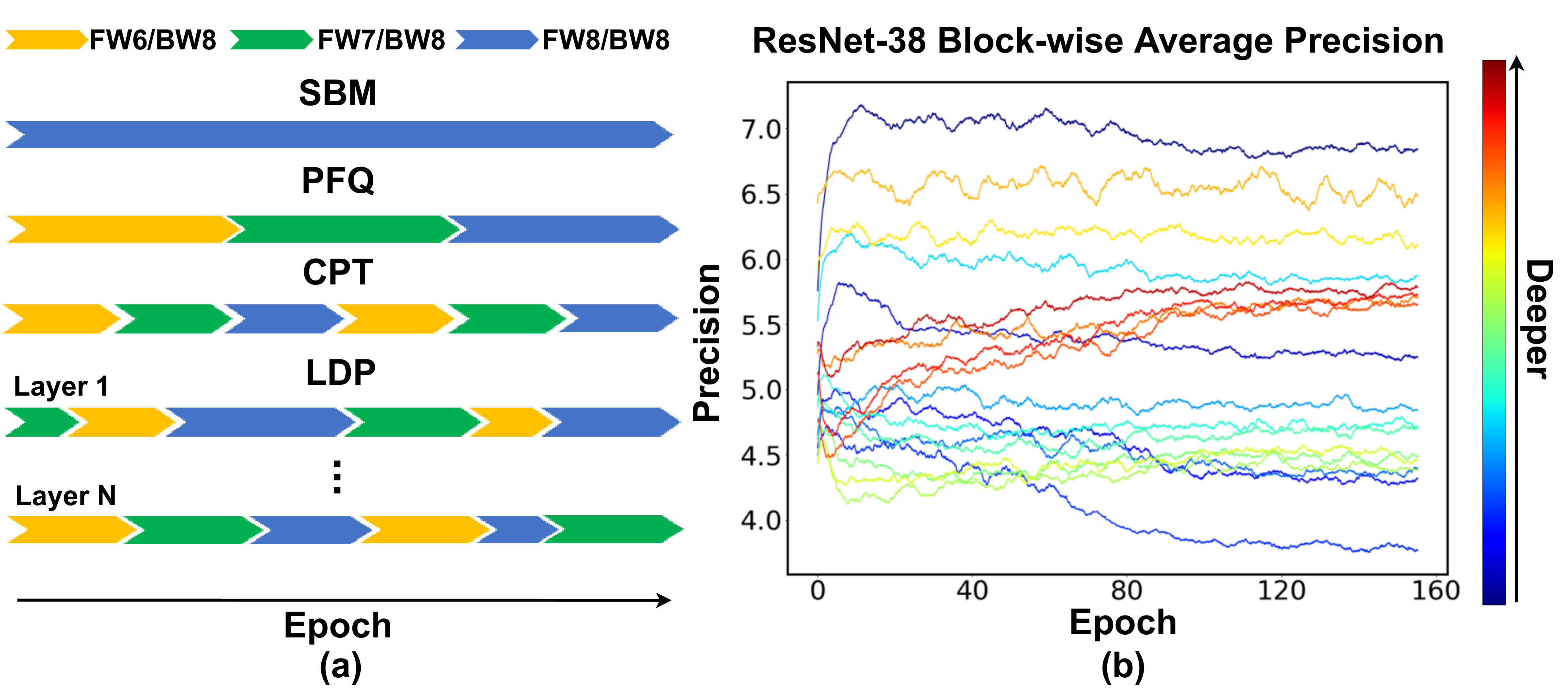}
    }
    \vspace{-1.5em}
    % \caption{(a) A conceptual view with our proposed \textbf{L}earnable \textbf{D}ynamic \textbf{P}recision DNN training framework vs. (1) static low-precision training SBM~\cite{banner2018scalable}, dynamic low-precision training with PFQ~\cite{fu2020fractrain}, and CPT~\cite{fu2021cpt}. In the sub-figure, each line represents a precision schedule in a layer, where LDP adopts a \textbf{learned layer-wise dynamic precision schedule} to optimally balance the training efficiency and performance trade-off; and (b) the learned spatial precision distribution and temporal schedule of ResNet-20@CIFAR-100, where different curves correspond to different blocks. }  
    % \caption{(a) A conceptual view of our proposed \textbf{L}earnable \textbf{D}ynamic \textbf{P}recision DNN training framework vs. (1) static low precision training (SBM~\cite{banner2018scalable}), dynamic low precision training (PFQ~\cite{fu2020fractrain} and CPT~\cite{fu2021cpt}). Here, each row shows the precision schedule for the whole model in the baselines, whereas LDP adopts a \textbf{learned layer-wise dynamic precision schedule} to optimally balance the training efficiency and accuracy trade-off; and (b) LDP's learned spatial precision distribution and temporal precision schedule of ResNet-38@CIFAR-100, where different curves correspond spatial precision distribution to different residual blocks. The fractional precision is due to the block-wise average and moving average among iterations for better visualization. }
    \caption{(a) A conceptual view of our proposed \textbf{L}earnable \textbf{D}ynamic \textbf{P}recision DNN training framework vs. (1) static low precision training (SBM~\cite{banner2018scalable}), (2) dynamic low precision training, (PFQ~\cite{fu2020fractrain} and (3) CPT~\cite{fu2021cpt}). Here, each row shows the precision schedule for the whole model of the baselines, where LDP adopts a \textbf{learned layer-wise dynamic precision schedule} to optimally balance the training efficiency and accuracy trade-off; and (b) LDP's learned spatial precision distribution and temporal precision schedule for ResNet-38@CIFAR-100, where different curves correspond to spatial precision distributions of different residual blocks andthe fractional precision is due to the block-wise average and moving average among iterations for better visualization. }
    \vspace{-1em}
    \label{fig:overview}
\end{figure}

While various low precision training techniques have been proposed to boost DNNs' training efficiency~\cite{banner2018scalable,zhou2016dorefa,yang2020training,sun2019hybrid}, most of these techniques adopt a fixed precision allocation strategy throughout the whole training process, leaving a large room for further squeezing out bit-wise savings. Motivated by the recent pioneering works, which advocate that (1) different DNN layers behave differently through the training process~\cite{zhang2019all, veit2016residual} and (2) different DNN training stages favor different training schemes~\cite{smith2017cyclical},
a few pioneering works~\cite{fu2020fractrain, rajagopal2020multi,fu2021cpt} have proposed to adopt dynamic training precision, which varies the precision spatially (e.g., layer-wise precision allocation) and temporally (e.g., different precision in different training epochs) and shows promising training efficiency and optimality over their static counterparts. However, existing dynamic low precision training methods rely on manually designed
dynamic precision schedules~\cite{fu2020fractrain, rajagopal2020multi, fu2021cpt}, thus making it challenging to be directly applied to new models/tasks and limiting their achievable training efficiency.

% Inspired by the prior arts, we target a more general dynamic DNN training scheme without introducing new hyper-parameters and additional manual fine-tuning and make the following contributions:  
Inspired by prior arts and motivated by their limitations, we target a general dynamic DNN training scheme without the necessity of manual fine-tuning and make the following contributions: 

\begin{itemize}
    % \item We propose LDP, a \textbf{L}earnable \textbf{D}ynamic \textbf{P}recision (LDP) training framework in which the layer-wise precision of DNN activations and weights are jointly learned and updated in a differentiable manner through the training process. LDP can automatically learn to spatially and temporally allocate the training cost, achieving better efficiency-performance trade-offs for both training and inference. 
    % \item We propose LDP, a \textbf{L}earnable \textbf{D}ynamic \textbf{P}recision (LDP) training framework in which the layer-wise precision of DNN activations and weights are jointly learned and updated in a differentiable manner through the training process. Hence, LDP automatically learns to spatially and temporally allocate the computational cost during training, boosting the efficiency-accuracy trade-offs for both training and inference. 
    
    \item We propose LDP, a \textbf{L}earnable \textbf{D}ynamic \textbf{P}recision (LDP) training framework to automatically learn to spatially and temporally allocate the computational cost during training by assigning different precision to different layers in each iteration, boosting the efficiency-accuracy trade-offs for both training and inference. 
    
    \item To enable an end-to-end training scheme of our LDP, we propose to  update the layer-wise precision in a differentiable manner by using a learnable quantization step, thus our LDP can jointly learn the layer-wise precision along with the model weights through the training process.

    % \item We visualize the temporal precision schedule and spatial precision distribution of LDP trained DNNs on different tasks to better understand the desired precision allocation strategy at different training stages and DNN layers both during and after training, respectively, from which the drawn insights could inspire new innovations for low precision training schemes.
    % \item We visualize the automatically learned temporal precision schedule and spatial precision distribution of LDP trained DNNs on different tasks in order to better understand the desired precision allocation strategy at \textit{different training stages and different DNN layers} for both \textit{training and inference}, respectively. The drawn insights could inspire future innovations for low precision training schemes.
    
    % \item Extensive experiments and ablation studies on seven networks, five datasets, and three tasks validate that LDP consistently achieves better efficiency-performance trade-offs over the state-of-the-art (SOTA) low precision training methods for both training and inference. Specifically, compared with the best SOTA baseline, LDP achieves a 0.31\% higher accuracy with a 39.1\% FLOPs reduction during training, when training ResNet-20 on CIFAR-10; and the LDP trained ResNet-74 directly leads to 49.8\% reduced cost during inference. 
    \item Extensive experiments and ablation studies on seven networks, five datasets, and three tasks validate that LDP consistently achieves better efficiency-accuracy trade-offs over state-of-the-art (SOTA) low precision training methods for both training and inference. Specifically, compared with the best SOTA baseline, LDP achieves a 0.31\% higher accuracy with a 39.1\% FLOPs reduction during training, when training ResNet-20 on CIFAR-10; and the LDP trained ResNet-74 directly leads to a 49.8\% reduced cost during inference. 
\end{itemize}

\section{Related Works}
% \textbf{DNN quantization.}
% DNN quantization~\cite{wu2018deep,han2015deep,zhang2018lq,faraone2018syq,yu2020kernel,zhou2017incremental,choi2018pact, banner2018scalable, zhou2016dorefa} is a well established DNN compression technique that aims to reduce the complexity of DNNs from the finest bit-level, achieving better efficiency-performance trade-offs. 
% % For example, \cite{zhou2017incremental} proposes to gradually quantize pretrained DNNs to achieve minimal accuracy degradation; \cite{faraone2018syq} proposes to learn a symmetric quantization codebook to alleviate the information loss under extremely low precisions; \cite{choi2018pact, elthakeb2020releq, esser2019learned} introduce learnable quantizers to boost the accuracy of extremely low precision DNNs; 
% Most existing works adopt a fixed precision for all layers~\cite{zhou2017incremental, faraone2018syq, choi2018pact, elthakeb2020releq}.
% Considering layer-wise difference in DNNs, \cite{wang2019haq, elthakeb2020releq, song2020drq} assign different precisions for different layers for better efficiency-performance trade-offs. 
% Despite the recent trending in mixed-precision DNNs, it is still unexplored regarding how to determine the layer-wise precision in each iteration during training, while it is computationally prohibitive to adopt expensive search space exploration methods like~\cite{wang2019haq} in each iteration. Our LDP tackles this via jointly learning the layer-wise precision and model weights to automatically determine the best training precision in each iteration.
\textbf{DNN quantization.}
DNN quantization~\cite{wu2018deep,han2015deep,zhang2018lq,faraone2018syq,yu2020kernel,zhou2017incremental,choi2018pact, banner2018scalable, zhou2016dorefa} is a popular DNN compression technique that aims to reduce the complexity of DNNs from the finest bit-level for achieving better efficiency-performance trade-offs. 
Most existing works adopt a fixed precision for all layers~\cite{zhou2017incremental, faraone2018syq, choi2018pact, elthakeb2020releq}.
Considering the layer-wise difference in DNNs, \cite{wang2019haq, elthakeb2020releq, song2020drq} assign different precisions for different layers during inference, leading to better efficiency-performance trade-offs. 
Despite the recent trend in mixed-precision DNNs, it is still underexplored regarding how to determine the layer-wise precision in each iteration during training. It is computationally prohibitive to do so in each iteration by adopting expensive search space exploration methods like~\cite{wang2019haq} did for inference. LDP tackles this via jointly learning the layer-wise precision and model weights to automatically determine the best training precision allocation in each iteration.

% \begin{figure}
%     \centering
%     \resizebox{\linewidth}{!}{
%     \includegraphics{Figures/LDP overview.pdf}
%     }
%     \caption{The proposed LDP DNN training framework. }
%     \vspace{-2em}
%     \label{fig:framework}
% \end{figure}
\begin{figure}
    \centering
    \resizebox{\linewidth}{!}{
    \includegraphics{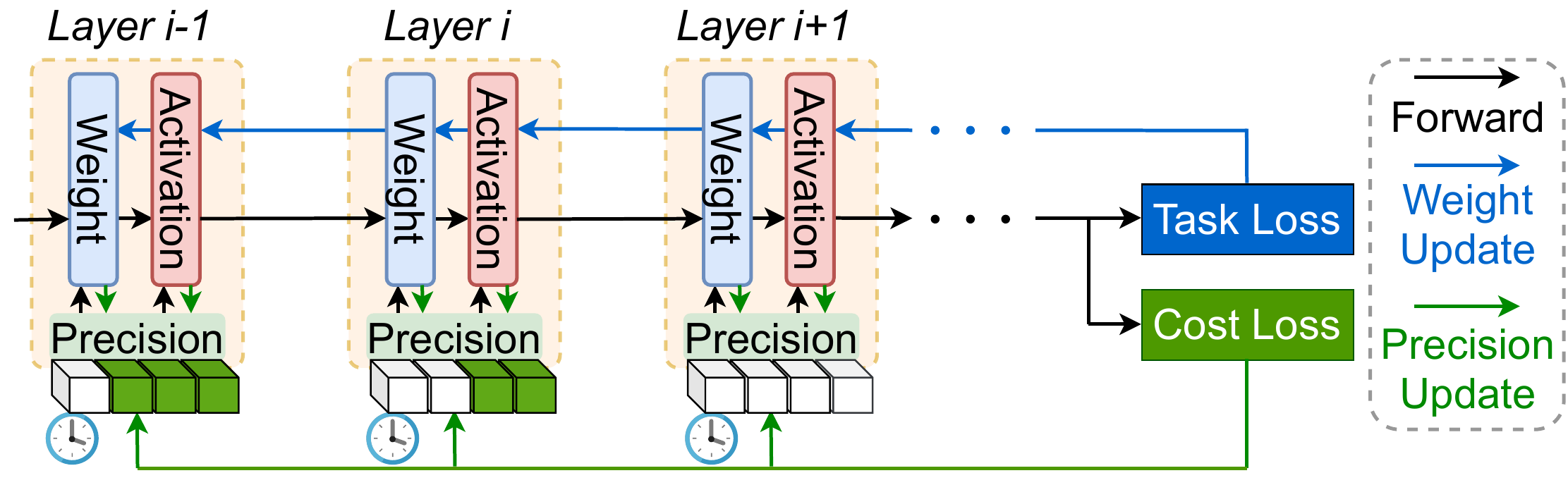}
    }
    \vspace{-2em}
    \caption{The proposed LDP training framework for DNNs. }
    \vspace{-1em}
    \label{fig:framework}
\end{figure}

% \textbf{Low precision training.}
% Pioneering works~\cite{banner2018scalable, yang2020training, zhou2016dorefa,sun2019hybrid, gupta2015deep} have shown that DNNs can be trained with a reduced precision without accuracy degradation. 
% There are two major application scenarios for low precision training: (1) for reducing communication cost instead of computational cost in distributed learning, \cite{wen2017terngrad,bernstein2018signsgd}; and (2) for efficient on-device/centralized learning, \cite{banner2018scalable, zhou2016dorefa, sun2019hybrid} use a reduced precision to achieve a comparable accuracy at reduced training costs. However, most of these methods adopt a pre-defined fixed precision for all layers through the training process. 
% Given the difference among DNN layers~\cite{zhang2019all}, those methods lack the flexibility of selecting the optimal precision patterns for different layers during different training stages. 
% A few pioneering works~\cite{fu2020fractrain, rajagopal2020multi} gradually vary the precision during training, and~\cite{fu2021cpt} proposes cyclical precision training, which yet requires manually designed dynamic precision schedules.
% In contrast, LDP can automatically learn a dynamic precision schedule both temporally in each iteration and spatially for each layer during training without the need for manual tuning.
\textbf{Low precision training.}
Pioneering works~\cite{banner2018scalable, yang2020training, zhou2016dorefa,sun2019hybrid, gupta2015deep} have shown that DNNs can be trained with a reduced precision without accuracy degradation. 
There are two major motivating applications for low precision training: (1) reducing communication cost instead of computational cost in distributed learning~\cite{wen2017terngrad,bernstein2018signsgd}; and (2) efficient on-device/centralized learning, e.g., \cite{banner2018scalable, zhou2016dorefa, sun2019hybrid} use a reduced precision to achieve a comparable accuracy at reduced training costs. However, most existing methods adopt a pre-defined fixed precision for all layers through the training process. 
Given the difference among DNN layers~\cite{zhang2019all}, those methods lack the flexibility of selecting the optimal precision patterns for different layers during different training stages. 
A few pioneering works \cite{wang2020dual} develops a sign prediction for low-cost, low-precision back-propagation during DNN training,
~\cite{fu2020fractrain, rajagopal2020multi} propose to gradually vary the precision during training, and~\cite{fu2021cpt} proposes the cyclical precision training, which yet requires manually designed dynamic precision schedules.
In contrast, LDP can automatically learn a dynamic precision schedule both temporally in each iteration and spatially for each layer during training without the need for manual tuning.

% \textbf{Dynamic precision DNNs.}
% Dynamic precision DNNs have mostly been discussed for efficient inference, which aim to dynamically allocate the precision throughout a network to achieve a higher inference efficiency, often at the cost of a higher training cost. 
% For example,~\cite{shen2020fractional} proposes to use a gating function to adapt the layer-wise precision in an input dependent manner, while \cite{yang2020fracbits} proposes to use linear interpolation to learn a fractional precision of each layer/filter;
% \cite{zhuang2018towards} proposes to start from a pretrained DNN and then gradually decrease the model precision to the target one. 
% All these methods incur additional training costs.  
% % Specifically, \cite{shen2020fractional} requires to additionally train the gating function, \cite{yang2020fracbits} needs a two-times higher computational cost to perform the interpolation between two precisions, and \cite{zhuang2018towards} needs additional training on top of the pretrained DNNs.   
% In parallel, techniques that achieve both efficient training and inference are highly desired, motivating our LDP framework.
\textbf{Dynamic precision DNNs.}
Dynamic precision DNNs have mostly been discussed for efficient inference \cite{wang2020dual,shen2020fractional}. They aim to dynamically allocate the precision throughout a DNN to achieve a higher inference efficiency, which often come with a higher training cost. 
For example,~\cite{shen2020fractional} proposes to use a gating function to adapt the layer-wise precision in an input dependent manner, while \cite{yang2020fracbits} proposes to use linear interpolation to learn a fractional precision of each layer/filter; and
\cite{zhuang2018towards} proposes to start from a pretrained DNN and then gradually decrease the model precision to the target one. 
All these methods incur additional training costs.  
In parallel, techniques that achieve both efficient training and inference are highly desired, motivating our LDP framework.

\section{The Proposed LDP Framework}

\begin{table*}[htb]
    \centering
        % \caption{The test accuracy of ResNet-38 trained on CIFAR-100 with \textbf{different layer-wise precision schedules} through the training process. In each training stage, $[a,b,c]$ represents assigning $a$-bit, $b$-bit, and $c$-bit to the first three blocks, respectively.}
       \caption{The test accuracy of ResNet-38 trained on CIFAR-100 with \textbf{different layer-wise precision schedules} through the training process. In each training stage, $[a,b,c]$ represents assigning $a$-bit, $b$-bit, and $c$-bit to ResNet-38's first three blocks, respectively.}
        % $a$-bit to the first block, $b$-bit to the second block, and $c$-bit to the third block in ResNet-38 at the corresponding stage, respectively. We use the number of Giga bit operations (GBitOPs) to measure the training cost. Note that all the three cases have exactly the same training cost. }
        \vspace{-0.2em}
        \resizebox{0.8\textwidth}{!}
        {
        \begin{tabular}{cccccc}
        \toprule[1pt]
        \multicolumn{4}{c}{Training Stages} & \multirow{2}{*}{Savings over static (\%)} & \multirow{2}{*}{Accuracy/\% } \\
        $[0\text{-th,} 30\text{-th}]$ & $[30\text{-th,} 60\text{-th}]$ & $[60\text{-th,} 90\text{-th}]$ & $[90\text{-th,} 160\text{-th}]$ \\
        \hline
        $[4,6,8]$ & $[6,8,4]$ & $[8,4,6]$ & $[8,8,8]$ & $1.10\times 10^8$ & $68.88 \pm 0.21 $ \\ 
        $[6,8,4]$ & $[8,4,6]$ & $[4,6,8]$ & $[8,8,8]$ & $1.10\times 10^8$ & $69.63\pm 0.14$ \\ 
        $[8,4,6]$ & $[4,6,8]$ & $[6,8,4]$ & $[8,8,8]$ & $1.10\times 10^8$ & $69.36 \pm 0.16$ \\ 
         \bottomrule[1pt]
         \end{tabular}
    }
    \vspace{-0.3em}
    \label{tab:hyop-layer}
\end{table*}

\begin{table}[htb]
    \centering
    % \caption{The test accuracy of ResNet-38 on CIFAR-100 with \textbf{different random precision change frequencies}. }
    \caption{The test accuracy of ResNet-38 on CIFAR-100 with \textbf{different frequencies for random precision change}. }
    \vspace{-0.2em}
    \resizebox{\linewidth}{!}{
    \begin{tabular}{c|cccc}
    \toprule[1pt]
        $k$ & 1 & 10 & 100 & Static\\
        \hline
        Accuracy/\% & $68.27 \pm 0.32$ & $69.71 \pm 0.26$ & $69.40 \pm 0.24$ & $69.62 \pm 0.10$ \\ 
        Training Cost/ & \multirow{2}{*}{$1.10\times 10^8$} & \multirow{2}{*}{$1.10\times 10^8$} & \multirow{2}{*}{$1.07\times 10^8$} & \multirow{2}{*}{$1.32\times 10^8$} \\
        GBitOPs &  &  &  & \\
        \bottomrule[1pt]
        \end{tabular}
        }
    \vspace{-1em}
    \label{tab:hypo-freq}
\end{table}
In this section, we first present the motivation and hypothesis inspiring our development of the LDP framework, and then describe LDP's design details.

\subsection{Motivating Observations}
\label{sec:hypo}
% \textbf{Key insight: The optimal layer-wise precision distribution varies in different training stages and it is challenging to determine the optimal spatial/temporal precision allocation strategy on-the-fly.}
% Existing works show that different DNN layers have different levels of sensitivity during training~\cite{zhang2019all} and different parts of DNNs respond differently to the same input~\cite{veit2016residual, greff2016highway}, suggesting that dynamic precision training can boost training efficiency without hurting the performance. 
% And~\cite{fu2021cpt} discovers that precision has a similar effect as the learning rate during training, providing a new angle to control the training process. 
% These findings suggest that different DNN layers require different precision schedules during training. Therefore, identifying the optimal precision allocation is the key to optimizing the efficiency-performance trade-offs. However, it is challenging to decide the best spatial/temporal precision allocation strategy during training in the huge search space as shown in following preliminary experiments.
\textbf{Key insights: The optimal layer-wise precision distribution varies in different training stages and it is challenging to determine the optimal spatial/temporal precision allocation strategy on-the-fly of training.}
Existing works show that different DNN layers have different levels of sensitivity during training~\cite{zhang2019all} and different parts of DNNs respond differently to the same inputs~\cite{veit2016residual, greff2016highway, wang2020dual}, suggesting that dynamic precision training can boost training efficiency without hurting the accuracy. 
Meanwhile,~\cite{fu2021cpt} discovers that precision has a similar effect as the learning rate during training, providing a new angle to control the training process. 
These findings suggest that different DNN layers require different precision schedules during training. Therefore, identifying the optimal precision allocation can further optimize the efficiency-accuracy trade-offs. However, it is challenging to decide the best spatial/temporal precision allocation strategy during training due to the huge search space as shown in the following experiments.

\begin{table*}[hbt]
    \centering
    %   \caption{The test accuracy, computational cost, and trained model's inference cost of ResNet-20/38/74 on CIFAR-10/100}
    \caption{The test accuracy, computational cost, and trained models' inference cost of ResNet-20/38/74 on CIFAR-10/100.}
    \vspace{-0.5em} 
    \resizebox{\textwidth}{!}{
    \begin{tabular}{ccccccccc}
        \toprule[2pt]
         \multicolumn{3}{c}{\textbf{Datasets}} & \multicolumn{3}{c}{\textbf{CIFAR-100}} & \multicolumn{3}{c}{\textbf{CIFAR-10}} \\
            \midrule
            Model & Method & Precision & Acc(\%) & Training Cost(GBitOps) & Inference Cost(GBitOps) & Acc(\%) & Training Cost(GBitOps) & Inference Cost(GBitOps) \\
            \midrule
            \multirow{10}{*}{ResNet-20} 
            & SBM & FW8/BW8 & 67.24 & 0.62e8 & 1.31 & 91.86 & 0.62e8 & 1.31\\
            & PFQ & FW3-8/BW8 & 67.31 & 0.50e8 & 1.31 &  91.75 & 0.50e8 & 1.31 \\
            & LDP & FW3-8/BW8 & \textbf{67.88} & \textbf{0.41e8}  & \textbf{0.71} & \textbf{92.08} & \textbf{0.41e8} & \textbf{0.70} \\
             \cmidrule{2-9}
            & Improv. &  & \textbf{+0.57} & \textbf{-18.0\%} & \textbf{-45.8\%} & \textbf{+0.22} & \textbf{-33.9\%} & \textbf{-46.6\%} \\
          \cmidrule[2pt]{2-9}
            & SBM & FW8/BW8 & 67.24 & 0.62e8 & 1.31 & 91.86 & 0.62e8 & 1.31\\
            & PFQ & FW4-8/BW8 & 67.47 & 0.51e8 & 1.31 & 91.66 & 0.50e8 & 1.31  \\
            & LDP & FW4-8/BW8 & \textbf{67.64} & \textbf{0.41e8} & \textbf{0.66} & \textbf{91.86} & \textbf{0.41e8} & \textbf{0.70} \\
            \cmidrule{2-9}
            & Improv. &  & \textbf{+0.17} & \textbf{-19.6\%} & \textbf{-49.6\%} & \textbf{+0.00} & \textbf{-33.9\%} & \textbf{-46.6\%} \\
          \midrule[2pt]
           \multirow{10}{*}{ResNet-38} 
            & SBM & FW8/BW8 & 69.38 & 1.33e8 & 2.69 & 92.69 & 1.33e8 & 2.69\\
            & PFQ & FW3-8/BW8 & 69.50 & 1.04e8 & 2.69 & 92.55 & 1.05e8 & 2.69  \\
            & LDP & FW3-8/BW8 & \textbf{69.77} & \textbf{0.87e8} & \textbf{1.35} & \textbf{92.73} & \textbf{0.86e8} & \textbf{1.36} \\
             \cmidrule{2-9}
            & Improv. &  & \textbf{+0.27} & \textbf{-16.3\%} & \textbf{-49.8\%} & \textbf{+0.04} & \textbf{-35.3\%} & \textbf{-49.4\%} \\
           \cmidrule[2pt]{2-9}
            & SBM & FW8/BW8 & 69.38 & 1.33e8 & 2.69 & 92.69 & 1.33e8 & 2.69\\
            & PFQ & FW4-8/BW8 & 69.72 & 1.07e8 & 2.69 & \textbf{92.70} & \textbf{1.08e8} & \textbf{2.69} \\
            & LDP & FW4-8/BW8 & \textbf{69.81} & \textbf{0.87e8} & \textbf{1.33} & 92.69 & 0.86e8 & 1.37 \\
            \cmidrule{2-9}
            & Improv. &  & \textbf{+0.09} & \textbf{-18.7\%} & \textbf{-50.6\%} & \textbf{-0.01} & \textbf{-20.4\%} & \textbf{-49.1\%} \\
           \midrule[2pt]
           \multirow{10}{*}{ResNet-74} 
            & SBM & FW8/BW8 & 71.05 & 2.67e8 & 5.42 & 93.30 & 2.67e8 & 5.42\\
            & PFQ & FW3-8/BW8 & 71.07 & 2.03e8 & 5.42 & 92.74 & 2.11e8 & 5.42 \\
            & LDP & FW3-8/BW8 & \textbf{71.28} & \textbf{1.72e8} & \textbf{2.83} & \textbf{93.63} & \textbf{1.72e8} & 2.82 \\
             \cmidrule{2-9}
            & Improv. &  & \textbf{+0.21} & \textbf{-15.3\%} & \textbf{-47.8\%} & \textbf{+0.33} & \textbf{-35.6\%} & \textbf{-48.0\%} \\
           \cmidrule[2pt]{2-9}
            & SBM & FW8/BW8 & 71.05 & 2.67e8 & 5.42 & 93.30 & 2.67e8 & 5.42\\
            & PFQ & FW4-8/BW8 & 71.15 & 2.16e8 & 5.42 & 93.45 & 2.21e8 & 5.42  \\
            & LDP & FW4-8/BW8 & \textbf{71.21} & \textbf{1.72e8} & \textbf{2.78} & \textbf{93.50} & \textbf{1.73e8} & \textbf{2.80} \\
            \cmidrule{2-9}
            & Improv. &  & \textbf{+0.06} & \textbf{-20.4\%} & \textbf{-48.7\%} & \textbf{-0.05} & \textbf{-21.7\%} & \textbf{-48.3\%} \\
        \bottomrule[2pt]
    \end{tabular}

    }
    \label{tab:cifar}
    \vspace{-1em}
\end{table*}

\textbf{Preliminary quantitative evaluation.} 
% \uline{Settings:} In Tab.~\ref{tab:hyop-layer}, we train ResNet-38 on CIFAR-100 for 160 epochs following the training setting in~\cite{wang2018skipnet}. 
% In particular, to \textbf{evaluate the impact of layer-wise precision schedules}, we divide the training into four stages: [0-th, 30-th], [30-th, 60-th], [60-th, 90-th], and [90-th, 160-th], and assign different precisions to different blocks of ResNet-38 in the first three training stages and resume a static 8-bit low precision in the last training stage, to evaluate the impact of assigning different precisions to different layers during training under the same total training GBitOPs budget. To \textbf{evaluate how the precision changing frequency affects the trained DNN performance}, in Tab.~\ref{tab:hypo-freq}, we randomly assign a precision value from $[4,6,8]$-bit to all layers of DNNs every $k$ iterations to quantize the DNN weights, activations, and gradients in the first 90 training epochs. 
% We evaluate experiments with different $k$ values in $[1, 10, 100]$.
\uline{Settings:} We conduct two experiments to (1) \textbf{evaluate the impact of layer-wise precision schedules} in Table~\ref{tab:hyop-layer}, and (2) \textbf{evaluate how the precision change frequency affects the trained DNN's accuracy} in Table~\ref{tab:hypo-freq}. We train ResNet-38 on CIFAR-100 for 160 epochs following the training setting in~\cite{wang2018skipnet} for both experiments. Specifically, in Table~\ref{tab:hyop-layer}, we divide the training into four stages: [0-th, 30-th], [30-th, 60-th], [60-th, 90-th], and [90-th, 160-th], and assign different precisions to different blocks of ResNet-38 in the first three training stages and adopt a static 8-bit low precision in the last training stage, in order to evaluate the impact of assigning different precisions to different layers during training under the same total training budget of GBitOPs (Gigabit operations). In Table~\ref{tab:hypo-freq}, we randomly assign a precision value from $[4,6,8]$-bit to all layers of the DNNs every $k$ iterations to quantize the DNN weights, activations, and gradients in the first 90 training epochs. In this experiment,
we evaluate results with different $k$ values in $[1, 10, 100]$. We report the average accuracy and the standard deviation of three runs for all experiments above. 

% \uline{Results:} We observe that (1) in Tab.~\ref{tab:hyop-layer}, different precision schedules through the training process vary the final accuracy by as high as $0.62\%$ under the same total training cost;
% % , and the precision schedules that allocate more computational budget (i.e., assign higher precisions) to the shallower layers in the early stages and gradually increase the computational budget to the deeper layers as the training process proceeds achieves higher accuracies; 
% and (2) in Tab.~\ref{tab:hypo-freq}, a good precision changing frequency leads to as high as a 0.53\% higher accuracy over other frequency and a $0.09\%$ higher accuracy with a $16\%$ lower training cost than the static precision training baseline.
\uline{Results:} We observe that (1) in Table~\ref{tab:hyop-layer}, different precision schedules through the training process vary the final accuracy by as high as $0.75\%$ under the same total training cost;
% , and the precision schedules that allocate more computational budget (i.e., assign higher precisions) to the shallower layers in the early stages and gradually increase the computational budget to the deeper layers as the training process proceeds achieves higher accuracies; 
and (2) in Table~\ref{tab:hypo-freq}, the best precision change frequency leads to (1) as high as a $1.44\%$ higher accuracy over other frequencies and (2) a $0.09\%$ higher accuracy with a $16\%$ lower training cost than the static precision training baseline.

% training with a random precision varied every $k=10$ iterations achieves the best accuracy, i.e., a 0.39\% and 0.53\% higher accuracy than that of the $k=1$ and $k=100$ cases under similar training costs, respectively, and a $0.09\%$ higher accuracy with a $16\%$ lower training cost than the static precision training baseline.     

% \uline{Analysis:} This set of experiments show that (1) given the same total training cost budget, how to allocate the training cost budget via spatially and temporally scheduling the training precision during training can greatly impact the finally achieved accuracy; (2) the precision changing frequency also impacts the achieved accuracy and even a naive randomly generated precision schedule with a properly selected precision changing frequency can offer a better training efficiency; and (3) no golden rule exists for determining the optimal spatial/temporal precision schedule, which highly relies on manual hyper-parameter tuning in SOTA methods~\cite{fu2020fractrain, rajagopal2020multi,fu2021cpt}, and it's challenging to automatically derive the optimal spatial/temporal schedule of training precision given the huge space of layer-wise precision schedule.
% \uline{Analysis:} This set of experiments show
\uline{Analysis:} This set of experiments shows that (1) given the same total training cost budget, how to allocate the training cost budget by spatially and temporally scheduling the training precision during training can significantly impact the finally achieved model accuracy; (2) the precision changing frequency also affects the achieved accuracy and even a naive randomly generated precision schedule with an adequately selected precision changing frequency can offer a better training efficiency; and (3) there exists no golden rule for determining the optimal spatial/temporal precision schedule, which highly relies on manual hyper-parameter tuning in SOTA methods~\cite{fu2020fractrain, rajagopal2020multi,fu2021cpt}, and it is challenging to automatically derive the optimal spatial/temporal schedule of training precision given the huge space of layer-wise precision schedule.

\subsection{The Proposed LDP Framework\label{sec:ldp}}
% Existing mixed-precision networks rely on costly trial-and-error methods (e.g., reinforcement learning-based~\cite{wang2019haq} and evolutionary-based~\cite{yuan2020evoq} ones) to determine the layer-wise precisions. 
% It is thus computationally impractical to apply these methods in each training step. 
% Therefore, we propose to make the precision aware of training states via jointly learning the layer-wise precision with the model weights in a differentiable manner. 

% As the precision itself is discrete and non-differentiable to the loss function, we introduce a continuous layer-wise learnable parameter $\beta^l$ for each layer $l$ with a quantization step size $s^l$ defined as:  
Existing mixed-precision networks rely on costly trial-and-error methods (e.g., reinforcement learning-based~\cite{wang2019haq} and evolutionary-based~\cite{yuan2020evoq} ones) to determine the layer-wise precision. 
It is thus computationally impractical to apply these methods in each training iteration. 
Therefore, we propose to make the precision be aware of training states via jointly learning the layer-wise precision with the model weights in a differentiable manner. 

As the precision itself is discrete and non-differentiable to the loss function, we introduce a continuous layer-wise learnable parameter $\beta^l$ for each layer $l$ with a quantization step size $s^l$ defined as:  

\begin{equation}
\label{eq:quant}
    s^l = \frac{R_{range}}{2^{\text{Round}(\beta^l\times N)} - 1}, 
\end{equation}
% \begin{equation}
%     s^l = \frac{R_{range}}{2^{N} - 1}, 
% \end{equation}
% \begin{equation}
%     s = \frac{R_{range}}{2^{N} - 1}, 
% \end{equation}
where $R_{range}$ is the dynamic range of input parameters, $N$ is the range of the available precision, and Round$(.)$ indicates rounding the value to the nearest integer. 
% Given a full precision value $I$ at layer $l$, its quantized counterpart $Q$ with $s^l$ can be defined as:
% \begin{equation}
%     Q = \text{Round}(\frac{I-I_{ZeroPoint}}{s^l})+ I_{ZeroPoint} ,
% \end{equation}
% where $I_{ZeroPoint}$ is an input-dependent parameter to normalize the inputs.  
% In this way, $\beta^l$ can be integrated into DNNs' computational flow and updated with respect to the loss function in a differentiable manner. where $R_{range}$ is the dynamic range of input parameters, $N$ is the range of the available precisions, and Round$(.)$ indicates rounding the value to the nearest integer. 
Given a full precision value $I$ at layer $l$, its quantized counterpart $Q$ with $s^l$ can be defined as:
\begin{equation}
    Q = \text{Round}(\frac{I-I_{ZeroPoint}}{s^l})+ I_{ZeroPoint} ,
\end{equation}
where $I_{ZeroPoint}$ is an input-dependent parameter for normalizing the inputs.  
In this way, $\beta^l$ can be integrated into DNNs' computational flow and updated with respect to the loss function in a differentiable manner.

% \textcolor{red}{Here is the typing eq.}
% \begin{equation}
%     A_{q}^{l+1} = \frac{ s_{A}^{l} \cdot s_{w}^{l} }{ s_{A}^{l+1} }A_{q}^{l} * W_{q}^{l}  
% \end{equation}
% \begin{equation}
%     \begin{aligned}
%         \frac{\partial{L}}{ \partial{\beta_{A}^{l}} }& = \frac{\partial{L}}{ \partial{A_{q}^{l+1}} } \cdot \frac{\partial{A_{q}^{l+1}}}{ \partial{\beta_{A}^{l}} } \\
%         & = \frac{\partial{L}}{ \partial{A_{q}^{l+1}} } \cdot \frac{s_{w}^{l}}{s_{A}^{l+1}} \cdot (\text{Round}(\frac{I}{s_{A}^{l}}) \cdot \frac{\partial{s_{A}^{l}}}{\partial{\beta_{A}^{l}}} \\
%         & + s_{A}^{l} \cdot \frac{\partial}{\partial{s_{A}^{l}}}(\frac{I}{s_{A}^{l}}) \cdot \frac{\partial{ s_{A}^{l} }}{\partial{ \beta_{A}^{l} }}) * W_{q}^{l}\\
%         & =  \frac{\partial{L}}{ \partial{A_{q}^{l+1}} } \cdot \frac{s_{w}^{l}}{s_{A}^{l+1}}\cdot (\text{Round}(\frac{I}{s_{A}^{l}}) - \frac{I}{s_{A}^{l}} ) \cdot \frac{\partial{s_{A}^{l}}}{\partial{\beta_{A}^{l}}} * W_{q}^{l} 
%     \end{aligned}
% \end{equation}
% \textbf{Effectively update $\beta$ and weight} 
% In the 

\textbf{Loss function.}
% As higher precisions favor more precise gradients and thus increased performance, directly updating the aforementioned $\beta^l$ in each layer $l$ with respect to only the task loss $L_{task}$ leads to a monotony increase of $\beta^l$ and thus a higher training cost. 
% This conflicts with the goal of LDP which is to learn a layer-wise dynamic precision schedule to better allocate the training cost within the network and during the training process. 
% To address this discrepancy, we incorporate the cost loss $L_{cost}$ into the network's loss function to control the balance between efficiency and accuracy, where
% $L_{cost}$ is defined as: 
% As higher precision
As a higher precision favors more precise gradients and thus increased accuracy, directly updating the aforementioned $\beta^l$ in each layer $l$ with respect to the task loss $L_{task}$ only leads to a monotony increase of $\beta^l$ and thus a higher training cost. 
This conflicts with the goal of LDP, which is to learn a layer-wise dynamic precision schedule to better allocate the training cost within the network and during the training process. 
To address this discrepancy, we incorporate a cost loss $L_{cost}$ into the network's loss function to control the trade-off balance between model efficiency and accuracy, where
$L_{cost}$ is defined as: 
\begin{equation}
L_{cost} = \left\{
             \begin{array}{lr}
             0,& \text{if }C < T \\
             C,& \text{if }C \geq T\\
             \end{array}
\right.
\end{equation}
where $T$ is the target iteration-wise training cost and $C$ is the forward pass cost in the current iteration defined as:  
\begin{equation}
    C = \sum_{l=1}^L O^l \times \frac{\text{Round}(\beta^l\times N)}{32}^2, 
\end{equation}
where $O^l$ is the required BitOPs for a full precision forward pass of layer $l$. However, the scale of $L_{task}$ and $L_{cost}$ can vary significantly throughout the training process and thus may require a tedious finetuning process to balance these two loss terms when applying LDP to different tasks. 
Thus, we adopt a balance factor $\alpha$ to balance the gradient of each layer's $\beta^l$ with respect to $L_{task}$ and $L_{cost}$. 
Specifically, the overall precision gradients $G^l$ for layer $l$ is defined as:
\begin{equation}
    \vspace{0.2em}
    G^l = G^l_{T} + \alpha \times G^l_{C} \times \frac{\text{Mean}(\text{Abs}(G_T))}{\text{Mean}(\text{Abs}(G_C)) + \epsilon},
    \vspace{0.4em}
\end{equation}
where $\text{Mean}(\text{Abs}(G_T))$ and $\text{Mean}(\text{Abs}(G_C))$ are the network-wise averaged absolute values for the precision gradients with respect to $L_{task}$ and $L_{cost}$, respectively, and $\epsilon$ is a small term to guarantee the training stability. 
To effectively prevent the precision from further growth, it is intuitive to constrain the contribution of $L_{task}$ and $L_{cost}$
% $L_{task}$'s and $L_{cost}$'s contribution
to $G^l$ to the same scale, 
so we set $\alpha=1$ in our implementation. 
In this way, when the precision is too high and the target training budget per iteration is exceeded, the cost term can effectively prevent the further increase in training precision without severely reducing the overall model precision, 
% avoiding the unrecoverable performance degradation. 
avoiding unrecoverable performance degradation. 
It is worth noting that although we calculate the gradient separately, it does not introduce additional computation costs. This is because $G^l_C$ can be naturally acquired once the network structure is fixed and does not need to specifically run backpropagation with respect to $L_{cost}$. 

\textbf{Potential hardware supports for LDP.} 
Scalable-precision architectures have been extensively studied~\cite{sharma2018bit, judd2016stripes, lee2018unpu} to support adaptive-precision execution of DNNs, i.e., select different precisions for different layers/iterations. In addition, it is promising to deploy LDP on other mixed-precision DNN accelerators~\cite{lee20197, kim20201b}.

\vspace{-1em}
\section{Experiments}
% In this section, we will first introduce the experiment setup, then benchmarking LDP results over SOTA training methods across various tasks, and finally 

\begin{figure}[tb]
    \centering
    % \vspace{-1em}
    \resizebox{\linewidth}{!}{
    \includegraphics{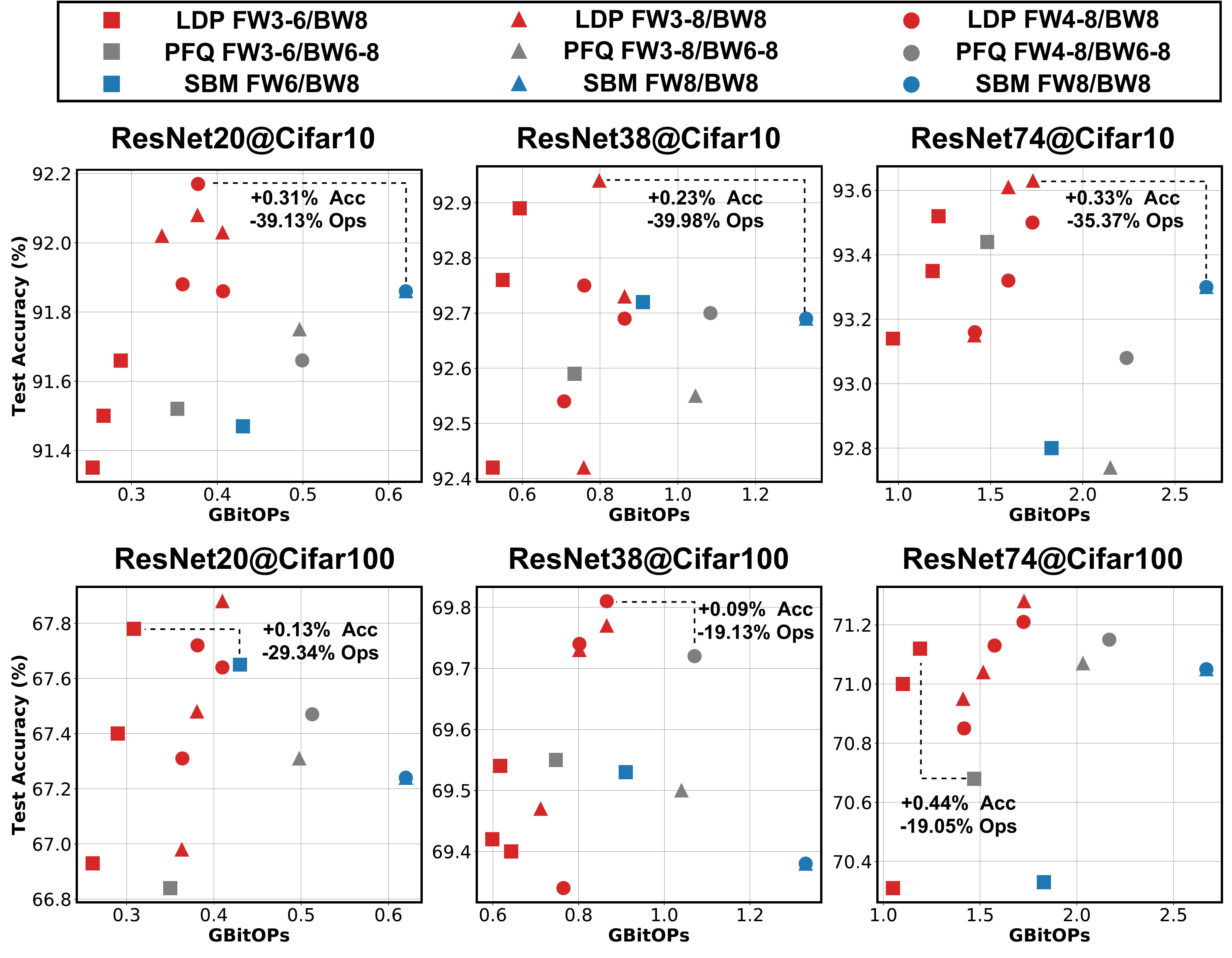}
    }
    \vspace{-2em}
    \caption{Benchmarking LDP with the SOTA static low precision training method SBM and dynamic precision training method PFQ in terms of model accuracy and training cost on ResNet-20/38/74 and CIFAR10/100.}
    \label{fig:cifar}
    \vspace{-1.5em}
\end{figure}

\subsection{Experiment Setup}
\textbf{Models, datasets, and baselines}. 
We evaluate our method on seven models (including four ResNet-based models~\cite{he2016deep}, Vision Transformer~\cite{touvron2021training}, Transformer~\cite{vaswani2017attention} and an efficient super-resolution model, PAN~\cite{zhao2020efficient}) and \uline{five  datasets} across \uline{three tasks} (including image classification on CIFAR-10/100~\cite{krizhevsky2009learning}, ImageNet~\cite{deng2009imagenet}, image super-resolution (SR) trained on DIV2K~\cite{agustsson2017ntire} and Flickr2K~\cite{timofte2017ntire} and evaluated on Urban-100~\cite{yang2010image}, and language modeling on WikiText-103~\cite{merity2017regularizing}). 
\uline{Baselines}: We benchmark the proposed LDP over SOTA low precision training methods, including PFQ~\cite{fu2020fractrain}, CPT~\cite{fu2021cpt}, and SBM~\cite{banner2018scalable}. For a fair comparison, we use the quantizer proposed in SBM~\cite{banner2018scalable} for all our baselines.

\textbf{Training settings}. We follow the standard training setting in all experiments, i.e., 
\cite{wang2018skipnet} and \cite{he2016deep} for CIFAR-10/100 and ImageNet, respectively, \cite{zhao2020efficient} for SR and \cite{baevski2018adaptive} for language modeling. Unless specifically specified, we use a learning rate of 0.1 and an SGD optimizer for learning the precision, i.e., $\beta^l$ in Eq.~\ref{eq:quant}, and we set $T$ as $60\%$ of the iteration-wise training cost $T_{stat}$ of the static low precision training baseline. For LDP, the training precision setting FW3-8/BW8 means the range of the learnable precision is 3$\sim$8-bit and the gradient is quantized to 8-bit; for PFQ and CPT, we follow the definition in their original paper~\cite{fu2020fractrain, fu2021cpt}.

% for classification tasks, we follow SOTA training settings in~\cite{wang2018skipnet} and~\cite{he2016deep} for CIFAR-10/100 and ImageNet, respectively. For SR and language modeling tasks, we follow the SOTA training setting in~\cite{zhao2020efficient} and \cite{baevski2018adaptive}, respectively. Unless otherwise specified, we use the learning rate of 0.1 and SGD optimizer for precision learning and we set $T$ as 60\% of the static low-precision training baseline's iteration-wise training cost $T_{stat}$. 

\subsection{Benchmark with SOTA Low Precision Training Methods}
\textbf{Benchmark on CIFAR-10/100.}
We first benchmark LDP with two SOTA low precision training methods: (1) the \textbf{static} low precision training method SBM~\cite{banner2018scalable}, and (2) the \textbf{dynamic} low precision training method PFQ~\cite{fu2020fractrain} on three different networks, i.e., ResNet-20/38/74, under two different precision schemes, i.e., FW3-8/BW8 and FW4-8/BW8 on CIFAR-10/100 datasets.
The results are shown in Table~\ref{tab:cifar}, where results with the highest accuracy are marked in bold. The accuracy improvement and training/inference cost reduction is \textbf{the difference between LDP and the strongest baseline with the highest accuracy under the same settings}. 
From Table~\ref{tab:cifar}, we have the following observations: (1) LDP consistently achieves better accuracy-training efficiency trade-offs than all baseline methods. Specifically, with $20.4\% \sim 39.6\%$ less training cost, LDP can achieve a comparable or even better accuracy ($-0.16\% \sim +0.56\%$) on CIFAR-10 and CIFAR-100 datasets; (2) LDP's learned precision naturally boosts the inference efficiency, reducing the inference cost by $29.0\% \sim 68.3\%$ compared with models trained with SBM or PFQ.
%  It is worth to notice that for more computational expensive models (e.g., ResNet-74), LDP achieves even higher inference cost reduction, making it easier to deploy powerful models trained with LDP on resource-constraint devices. 

To further evaluate the overall performance of LDP, we evaluate  LDP's performance under different $T\in[0.5T_{stat}, 0.7T_{stat}]$. The results are shown in Fig.~\ref{fig:cifar} and we can observe that: (1) the training cost can be effectively controlled by the value of $T$, indicating the easiness to fit LDP onto different training tasks with varied training budgets, and (2) LDP keeps achieving the best accuracy-efficiency trade-off under different training cost budgets. 
% Specifically, for ResNet-20 based on CIFAR-100 with FW3-8/BW8 precision setting, when varying $T\in[0.5T_{stat}, 0.7T_{stat}]$, the trained LDP consistently achieves over $0.5\%$ higher accuracy with the same or less training cost compared with the most competitive PFQ baseline. 

\begin{table}[tb]
    \centering
    \caption{The test accuracy, training cost, and trained models' inference cost of ResNet-18 and DeiT-Tiny on ImageNet.}
    \vspace{-1em}
    \resizebox{\linewidth}{!}{
    \begin{tabular}{cccccc}
    \toprule[2pt]
        \multirow{2}{*}{Model} & \multirow{2}{*}{Method} & \multirow{2}{*}{Precision} & \multirow{2}{*}{Acc(\%)} & Training Cost & Inference Cost \\
        &  &  &  & (GBitOps) & (GBitOps) \\
        \midrule
        \multirow{5}{*}{ResNet-18} & SBM & FW8/BW8 & 69.60 & 2.86e9 & 1.46e1  \\
        & CPT & FW4-8/BW8 & 69.64 & 1.99e9 & 1.46e1  \\
        & PFQ & FW4-8/BW6-8 & 69.12 & 2.47e9 & 1.46e1 \\
        & LDP & FW4-8/BW8 & 69.62 & 1.83e9  & 1.01e1  \\
         \cmidrule{2-6}
        & Improv. &  & -0.02 & -8.1\% & -30.8\% \\
        \midrule
        \multirow{5}{*}{DeiT-Tiny} & SBM & FW8/BW8     & 71.71 & 4.74e9 & 0.96e1 \\
           &  CPT & FW4-8/BW8   & 71.84 & 3.29e9 & 0.96e1 \\
           &  PFQ & FW4-8/BW6-8 & 71.70 & 3.96e9 & 0.96e1 \\
           &  LDP & FW4-8/BW8   & 71.92 & 3.08e9 & 0.67e1\\
        \cmidrule{2-6}
        & Improv. &  & +0.08 & -6.4\% & -30.2\%\\
        \bottomrule[2pt]
    \end{tabular}
    }
    \vspace{-1em}
    \label{tab:imagenet}
\end{table}

\textbf{Benchmark on ImageNet.}
We further verify the scalability of LDP on the more challenging ImageNet dataset across different model architectures. 
As shown in Table~\ref{tab:imagenet}, LDP still achieves comparable accuracy with less training cost compared with the most competitive baseline methods. 
Specifically, compared with CPT, LDP achieves a $0.08\%$ higher accuracy with $10.6\%$ less training cost to train DeiT-Tiny under FW3-8/BW8. Moreover, the LDP trained models are still more efficient than models trained with other baseline methods with an improvement in inference efficiency ranging between $11.0\% \sim 35.2\%$. 
% Setting:
% \begin{itemize}
%     \item Motivation: Similar to CIFAR, compare with widely used settings to prove the superiority in accuracy-efficiency trade-off of the proposed LDP.
%     \item Model: ResNet-18, ResNet-34, DeiT-Tiny, DeiT-Small 
%     \item Dataset: ImageNet 
%     \item Baseline: PFQ, CPT, SBM
%     \item Performance metric: Accuracy v.s. training cost in GFLOPs, (inference efficiency) 
%     \item results in Table~\ref{tab:imagenet}
% \end{itemize}

% \begin{itemize}
%     \item Motivation: Evaluate LDP on lower level vision tasks, which is a untouched area in previous low-precision training
%     \item Model: PAN 
%     \item Dataset: Evaluate on Urban-100, Set-14 and B-100
%     \item Baseline: SBM, CPT and PFQ
%     \item Performance metrics: PSNR, training cost in GFLOPs (and inference cost)
%     \item results in Table~\ref{tab:sr}
% \end{itemize}

\begin{table}[]
    \centering
    \caption{The PSNR and inference cost of PAN on Urban-100. }
    \vspace{-1em}
    \resizebox{\linewidth}{!}{
    \begin{tabular}{cccc}
    \toprule[2pt]
        \multirow{2}{*}{Method} & \multirow{2}{*}{Precision} & \multirow{2}{*}{Urban-100} & Inference Cost\\
         &  &  & (GBitOps) \\
        \midrule
        Half-Precision & FW16/BW16  & 26.01 & 6.43e1\\
        PFQ & FW8-16/BW16 & 25.99 & 6.43e1\\
        CPT & FW8-16/BW16 & 26.01 & 6.43e1\\
        LDP & FW8-16/BW16 & 26.03 & 5.22e1\\
        \midrule
        \multicolumn{2}{c}{Improv.} & +0.02 & -18.8\%\\
        % \midrule 
        % SBM & 8/8 &  &  &  &  & \\
        % PFQ & 3-8/8 &  &  &  &  & \\
        % CPT & 3-8/8 &  &  &  &  & \\
        % LDP & 3-8/8 &  &  &  &  & \\
        \bottomrule[2pt]
    \end{tabular}
    }
    \vspace{-1em}
    \label{tab:sr}
\end{table}

% \begin{itemize}
%     \item Motivation: Apply LDP to language models to show LDP can also be generalized to NLP tasks. 
%     \item Model: Transformer 
%     \item Dataset: WikiText-103
%     \item Baseline: SBM, PFQ, CPT
%     \item Performance metrics: Perplexity, Training cost in GFLOPs
%     \item precision: 3-8/8
%     \item results in Table~\ref{tab:nlp}
% \end{itemize}

\begin{table}[]
    \centering
    \caption{The test perplexity (the lower, the better) and training cost of Transformer on WikiText-103.}
     \vspace{-1em}
    \resizebox{\linewidth}{!}{
    \begin{tabular}{cccc}
    \toprule[2pt]
        Method & Precision & Perplexity & Training Cost (GBitOps) \\
        \midrule
        SBM & FW8/BW8 & 31.77 & 9.87e5\\
        % CPT & 4-8/8 & 30.22 & 7.66e5\\
        LDP & FW4-8/BW8 & 30.81 & 7.31e5\\
        \midrule
        \multicolumn{2}{c}{Improv.} & -0.96 & -25.9\% \\
        \bottomrule[2pt]
    \end{tabular}
    }
    \vspace{-1em}
    \label{tab:nlp}
\end{table}

\textbf{Benchmark on SR task.}
We also evaluate LDP on the SR task. It is noteworthy that given the nature of the relatively smaller gradient of the SR task, it is non-trivial to train the SR models with reduced precision. 
% To guarantee the training efficiency, we leave the first convolution layer and last upsampling layers unquantized and only quantize the feature extraction part in PAN with a precision scheme of FW8-16/BW16. 
We only quantize the feature extraction part in PAN with a precision scheme of FW8-16/BW16 and a precision learning rate of $20$ and $T=0.85T_{stat}$. 
The results are shown in Table~\ref{tab:sr} that LDP achieves a 0.02dB higher peak signal-to-noise ratio (PSNR) with $18.8\%$ less inference cost on Urban-100 compared with the original half-precision training, showing LDP's ability in further boosting models performance. 

\textbf{Benchmark on natural language processing (NLP) task.}
To validate the general effectiveness of LDP on NLP tasks, we also apply LDP on a language modeling task WikiText-103~\cite{merity2017regularizing} on top of the Transformer~\cite{vaswani2017attention} model. As shown in Table~\ref{tab:nlp}, LDP consistently achieves better performance than the SBM baseline with less training and inference cost, showing that the proposed LDP can be a general technique among different tasks.

\textbf{Ablation study about the choice of $\alpha$.}
To evaluate LDP's sensitivity to $\alpha$, we apply LDP on ResNet-20/38 with different values of $\alpha$. 
As shown in Table~\ref{tab:alpha}, different values of $\alpha$ do not have a significant impact on the achieved accuracy ($<0.2\%$ accuracy variance), suggesting the easiness of deploying LDP without the need for exhaustive hyperparameters finetuning.
% When $\alpha$ is too small, the high precision cannot be effectively penalized, leading to an inferior accuracy-training efficiency trade-off. 
% On the other hand, when $\alpha$ is too large, the large penalty applied to precision lead to slightly inferior performance. But the LDP's outstanding learning ability can still achieve comparable performance, suggesting the proposed LDP is not sensitive to the selection of $\alpha$. 

\subsection{Visualization of LDP's Learned Precision}
We visualize the temporal and spatial precision distribution learned by LDP to understand the characteristics of different models' learning process and the redundancy of different modules. 
% \textbf{Temporal precision distribution during training}
% \begin{itemize}
%     \item Motivation: Investigate how precision distribution of different models varies through training process. 
%     \item Setting: Visualize from trained models above 
%     \item Model: ResNet@CIFAR, PAN@Urban-100, DeiT@ImageNet, Transformer@NLP
%     \item Results in Fig.~\ref{fig:temporal_prec}
% \end{itemize}

\begin{table}[tb]
    \centering
    \caption{Training ResNet-20/38 on CIFAR-100 with different $\alpha$ using FW3-8/BW8.}
     \vspace{-1em}
    \resizebox{\linewidth}{!}{
    \begin{tabular}{c|cc|cc|cc}
    \toprule[2pt]
         & \multicolumn{2}{c|}{$\alpha=1.5$} & \multicolumn{2}{c|}{$\alpha=1$} & \multicolumn{2}{c}{$\alpha=0.5$} \\
         & Acc (\%) & GBitOps & Acc (\%) & GBitOps & Acc (\%) & GBitOps \\
         \midrule[2pt]
        ResNet-20 & 66.98 & 0.39e8 & 67.08 & 0.41e8 & 67.13 & 0.53e8\\
        ResNet-38 & 69.85 & 0.76e8 & 70.06 & 0.80e8 & 69.94 & 1.14e8\\
        \bottomrule[2pt]
    \end{tabular}
    }
    \label{tab:alpha}    
    \vspace{-1em}

\end{table}

\begin{figure}[tb]
    \vspace{-1em}
    \centering
    \includegraphics[width=\linewidth]{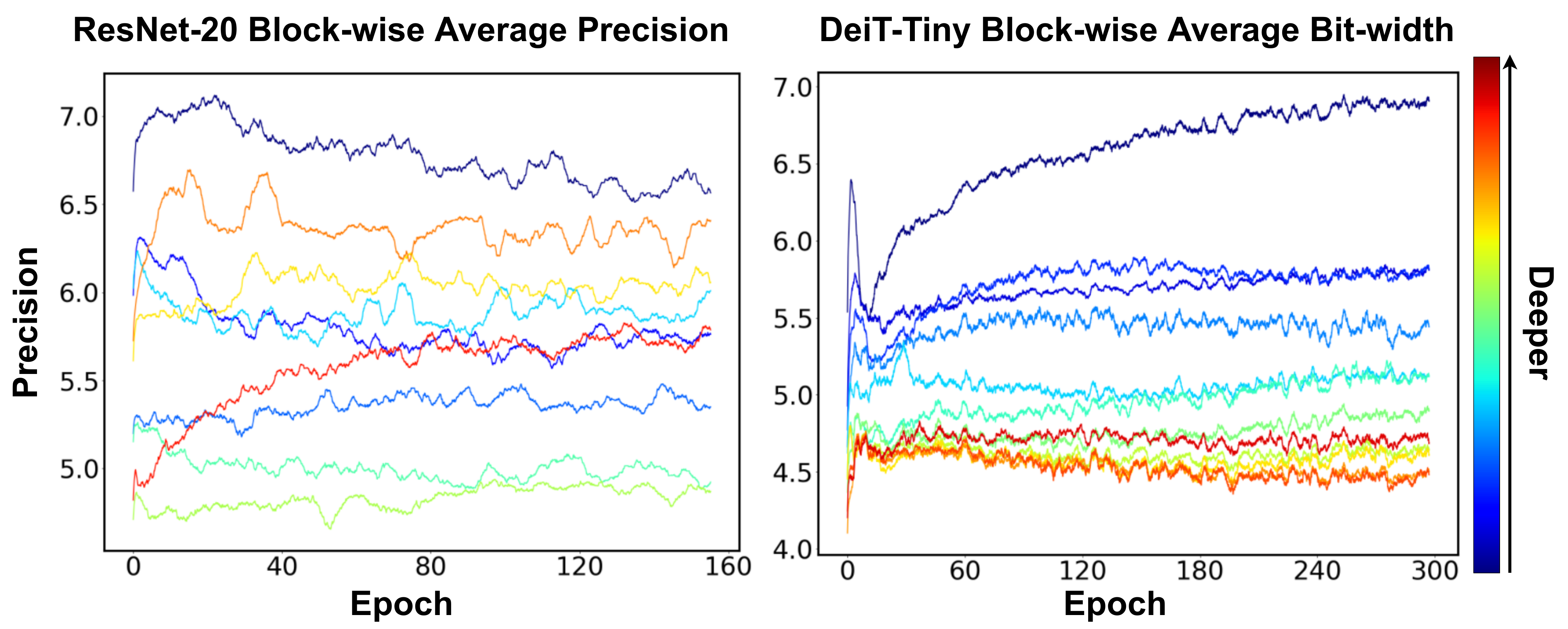}
    \vspace{-2.5em}
    \caption{Visualization of spatial precision distribution and temporal precision schedule during training. The warmer the line is, the deeper the corresponding layer is in the model. }
    \label{fig:temporal_prec}
    \vspace{-2em}
\end{figure}

\textbf{Temporary precision distribution during training.}
We first visualize the learned precision schedule of ResNet-20@CIFAR-100 and DeiT-Tiny@ImageNet through the training process. As shown in Fig.~\ref{fig:temporal_prec}, there are three patterns in the learned precision schedule of ResNet-20: (1) For shallower layers, the block-wise average precision first rapidly grows to high precision in the initial iterations of training, then gradually decrease, (2) for layers in the middle of the model, the precision keeps high through the whole training process, and (3) for deeper layers, the precision would gradually increase along with the training process. 

Such learned  precision schedule aligns with the understanding of ResNet's learning process~\cite{achille2018critical}, where high precision in shallow layers helps to learn the low-level information in the initial training stage; while at a later training stage, shallow layers' learned features can be represented in low precision and higher precision in deeper layers help better abstract semantic information from the low-level features passed from shallower layers. 
% Such learned precision schedule fits the understanding of ResNet's learning process. During the initial phase of training, the network first needs to have a relatively high precision schedule to learn to process the low-level pattern information and the deeper layers cannot effectively learn the distribution during the initial stage of training as the features learned by shallow layers are still changing without a stable distribution. 
% During the later phase of training, the pattern learned by shallow layers is basically fixed. Reducing the precision does not hurt the learned pattern. On the other hand, during this phase, deeper layers need higher precision to better learn the distribution from leaned features passed by shallower layers and thus need higher precision. 

On the other hand, the precision schedule in the DeiT-Tiny model is pretty different from that of ResNet-20 due to the difference in the model structure, leading to higher difficulty for training, as suggested in~\cite{touvron2021training, steiner2021train}. We notice the precision schedule of all blocks keeps growing to certain block-specific values. 
% The shallower the layer is, the higher this precision value is.
Such gradually increased spatial precision distribution echos the DeiT block-wise redundancy analyzed in~\cite{zhou2021deepvit}.
% showing LDP's ability in automatically learning the optimal precision schedule during training.

\begin{figure}
    \centering
    \vspace{-1em}
    \includegraphics[width=\linewidth]{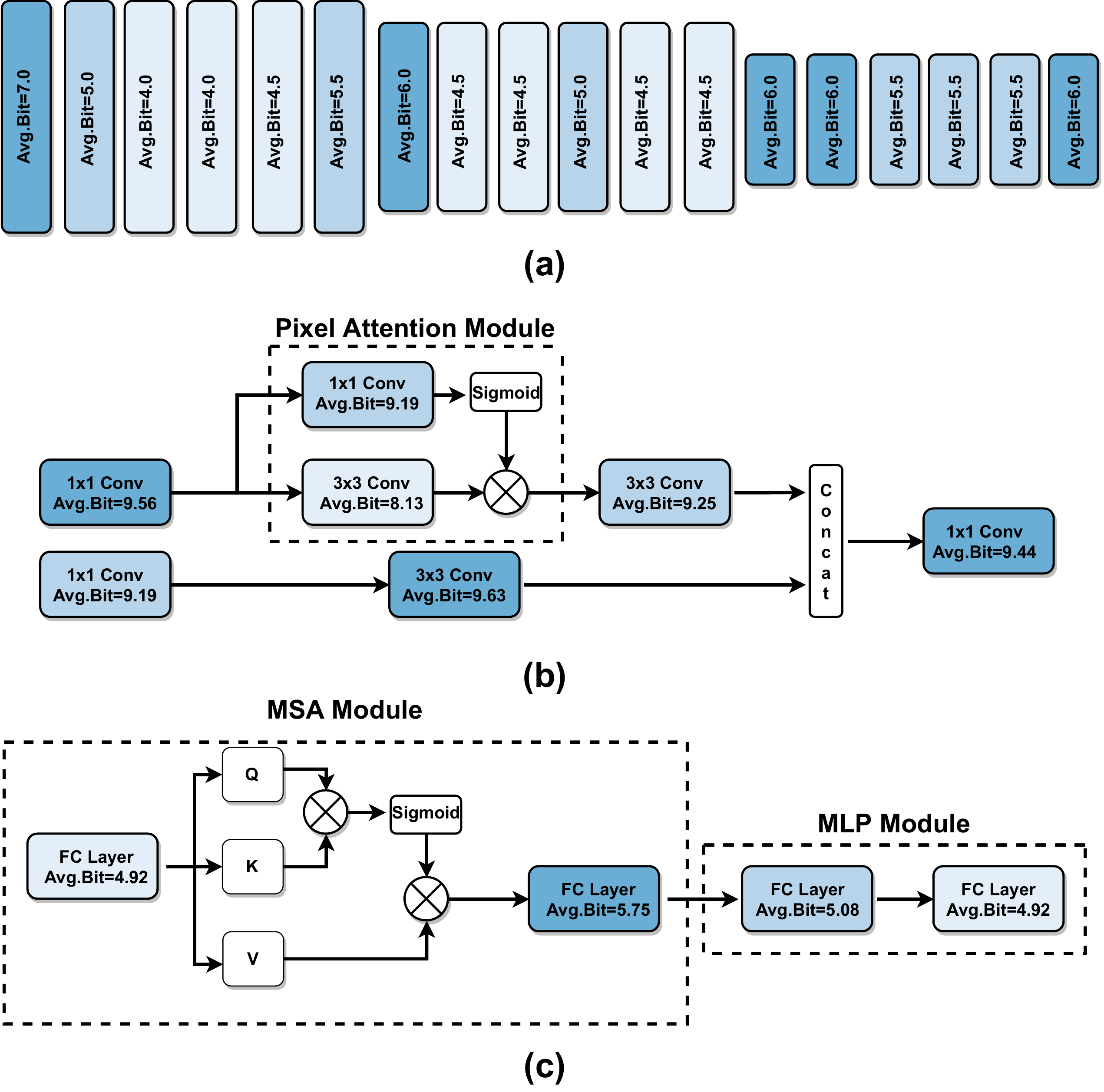}
    \vspace{-2.5em}
    \caption{Spatial precision distribution of models where \uline{deeper color indicates a higher precision}. (a) Block-wise average precision of ResNet-38, layer-wise average precision of (b) PAN's SCPA module, and (c) DeiT-Tiny's blocks. }
    \vspace{-1.5em}
    \label{fig:spatial}
\end{figure}

\textbf{Learned spatial precision distribution.}
We further investigate the learned spatial precision distribution on ResNet-38 (Fig.~\ref{fig:spatial}(a)), PAN (Fig.~\ref{fig:spatial}(b)) and DeiT-Tiny (Fig.~\ref{fig:spatial}(c)), respectively. 

In ResNet-38, LDP learns (1) significantly higher precisions in blocks containing downsampling layers, and (2) higher precision in the deep blocks with the lowest resolution, indicating less redundancy in these blocks which aligns with the observations in~\cite{wang2020dual, shen2020fractional}.

In the SCPA module in PAN, we observe that in the two-stream architecture, there exists a layer with relatively high precision in each branch of PA, suggesting that it is critical to preserve the detail of information in each branch.

We further visualize the DeiT block's spatial precision distribution and observe that the sequentially connected fully-connected (FC) layers have a gradually decreased precision, indicating the redundancy in the stacked FC layers. This motivates the necessity of reducing the redundancy in such layers, aligning with the observations in~\cite{guo2021cmt,park2022vision}.

\section{Conclusion}
% In this paper, we propose the \textbf{L}earnable \textbf{D}ynamic \textbf{P}recision training (\textbf{LDP}) for efficient and effective low precision training. By adding a learnable precision parameter, LDP can automatically learn a spatial precision distribution and a temporal precision schedule for each iteration on-the-fly during the training process. Extensive experiments, ablation studies and visualizations verify that LDP can learn the optimal precision schedule spatially and temporarily, achieving a better performance and training cost trade-off. 

In this paper, we propose a \textbf{L}earnable \textbf{D}ynamic \textbf{P}recision training framework called LDP for efficient and effective low precision training. By adding a learnable precision parameter, LDP can automatically learn a spatial precision distribution and a temporal precision schedule for each iteration on-the-fly during the training process. Extensive experiments, ablation studies, and visualizations verify that LDP can learn an effective precision schedule spatially and temporarily, pushing forward the frontier of the trade-off between task performances and training cost. 

\section*{Acknowledgements}
This work is supported by the National Science Foundation (NSF) through the MLWiNS program
(Award number: 2003137) and the RTML program (Award number: 1937592).

\bibliographystyle{ACM-Reference-Format}
\bibliography{main}
\end{document}